\documentclass[letterpaper, 10 pt, conference]{ieeeconf}  
\pdfoutput = 1

\IEEEoverridecommandlockouts                              

\usepackage[dvipdfmx]{graphicx}
\usepackage{sty/ynlab}
\usepackage{dsfont}
\usepackage{subcaption}

\newcommand{\argmin}{\mathop{\rm argmin}\limits}

\newcommand{\mytag}[1]{\texttt{#1}}

\def\eg{\textit{e.g}\onedot} 
\def\ie{\textit{i.e}\onedot}

\def\etal{\textit{et al}\onedot}

\def\eg{{\it e.g.}}

\def\ie{{\it i.e.}}
\def\etal{{\it et al. }}

\title{\LARGE \bf
Ground and Flight Locomotion for Two-Wheeled Drones\\via Model Predictive Path Integral Control*
}

\author{Gosuke Kojima$^{1}$, Kohei Honda$^{2}$, Satoshi Nakano$^{1}$, and Manabu Yamada$^{1}$
\thanks{*This work was supported by JSPS KAKENHI Grant Numbers JP23K13352, JP24K07540.}
\thanks{$^{1}$Gosuke Kojima, Satoshi Nakano, and Manabu Yamada are with the Department of Engineering,
        Nagoya Institute of Technology, Aichi 466-8555, Japan
         {\tt\small  g.kojima.703@stn.nitech.ac.jp,} {\tt\small\{nakano, yamada.manabu\}@nitech.ac.jp}}%
\thanks{$^{2}$Kohei Honda is with the Department of Mechanical System Engineering, Nagoya University,
        Aichi 464-8601, Japan
        {\tt\small honda.kohei.k3@f.mail.nagoya-u.ac.jp}}%
}

\begin{document}

\maketitle
\thispagestyle{empty}
\pagestyle{empty}

\begin{abstract}
        \looseness=-1
This paper presents a novel approach to motion planning for two-wheeled drones that can drive on the ground and fly in the air. 
Conventional methods for two-wheeled drone motion planning typically rely on gradient-based optimization and assume that obstacle shapes can be approximated by a differentiable form.
To overcome this limitation, we propose a motion planning method based on Model Predictive Path Integral (MPPI) control, enabling navigation through arbitrarily shaped obstacles by switching between driving and flight modes. 
To handle the instability and rapid solution changes caused by mode switching, our proposed method switches the control space and utilizes the auxiliary controller for MPPI.
Our simulation results demonstrate that the proposed method enables navigation in unstructured environments and achieves effective obstacle avoidance through mode switching.
\end{abstract}

\section{Introduction}
\label{sec:introduction}
A groundbreaking vehicle that seamlessly integrates robust terrestrial navigation with boundless aerial freedom promises to overcome the trade-off between energy efficiency and reachability. 
As illustrated in Fig.\ref{fig:two_wheel_drone_model}, two-wheeled drones are one of the promising vehicles, which are capable of both energy-efficient \emph{driving} on the ground and high mobility \emph{flight} in the air~\cite{kalantari2013,takahashi2015}.
However, despite their appeal, two-wheeled drones present significant control challenges due to discontinuous changes in dynamics and constraints between driving and flight modes.  
Our research interest lies in safely and efficiently navigating them by managing the mode switching in the motion planning scheme.

A common approach to motion planning for the two-wheeled drone is model predictive control (MPC), which minimizes a finite-horizon cost function while considering system dynamics and constraints. 
However, gradient-based MPC has the challenge of inability to handle non-differentiable costs~\cite{wu2023,zhang2023}.
This challenge poses difficulties in avoiding arbitrarily shaped obstacles and in considering the dynamics at the moment of contact with the ground.
In contrast, sample-based MPC, such as model predictive path integral control (MPPI)~\cite{williams2018}, has proven to be more practical for robotic motion planning because it can handle non-differentiable costs, constraints, and dynamics.
MPPI can minimize a non-differentiable cost function through stochastic, sample-based search performed in real-time.

\begin{figure}[t]
  \centering
  \includegraphics[width=0.95\linewidth]{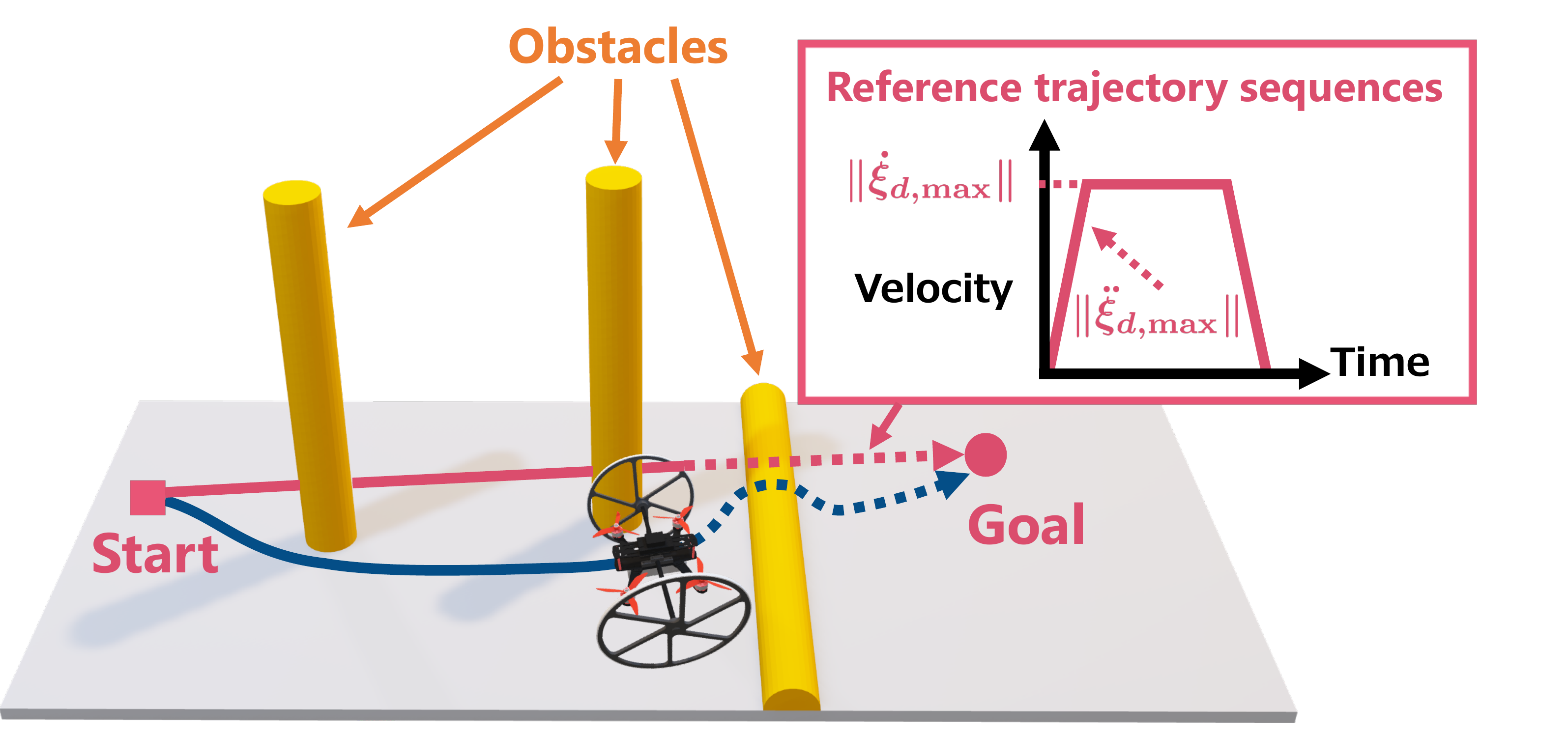}
  \caption{Our target scenario. A two-wheeled drone follows a reference trajectory while avoiding obstacles by switching between ground driving and flight modes.}
  \label{fig:problem_setting}
  \vspace{-6mm}
\end{figure}

\looseness=-1
In this work, we propose an MPPI-based motion planning method for two-wheeled drones, which can handle the switching between driving and flight modes with MPPI.
As a result, our method allows obstacle avoidance with arbitrary obstacles, a challenge to the prior methods~\cite{wu2023} by taking advantage of MPPI.
Furthermore, it can also take into account discontinuous contact dynamics in motion planning, which has been neglected by MPC-based methods~\cite{wu2023,zhang2023}.
To handle mode switching in MPPI, we propose an extension of MPPI that switches the input space and introduces an auxiliary controller for the solution search.
First, we switch the input space to suppress an unstable single-wheel landing near the ground.
Second, we design the prior distribution considering the multimodality of the two-wheeled drone to generate effective samples for solution search in switching modes.

We evaluate our method through a numerical simulation in an environment containing cylinder-shaped obstacles, as shown in Fig.~\ref{fig:problem_setting}.
Experimental results demonstrate that our method can navigate the two-wheeled drone by state-dependently switching between driving and flight modes while avoiding obstacles.
We also confirm that our proposed switching of the auxiliary controller for MPPI is effective in enhancing the efficiency of navigation.
\section{Related Work}
\label{sec:related_work}

\subsection{Motion Planning for Two-Wheeled Drones}
One approach to the control of a two-wheeled drone is a nonlinear cascade control with position and attitude controllers~\cite{colmenares-vazquez2019,pimentel2022,tang2024}, just as with normal drones~\cite{nonami2010}.
This approach enables trajectory tracking at low computational cost, but requires switching of controllers, which are individually designed for each mode, such as driving and flying.
These works also focus on trajectory tracking and do not consider obstacle avoidance, as we address in this paper.

\looseness=-1
Another approach employs gradient-based nonlinear MPC methods for trajectory tracking~\cite{zhang2023} and obstacle avoidance~\cite{wu2023}, which can incorporate mode switching into the optimization problem using complementarity constraints without preparing separate controllers for each mode.
However, the gradient-based MPC requires the optimization problem to be differentiable, which is challenging when handling non-differentiable costs and dynamics, such as onboard sensor information, \eg, cost maps~\cite{macenski2023}, and the momentary effects of repulsion with the ground.

\subsection{MPPI-based Motion Planning for Drones}
Several studies have proposed motion planning or control for drones using MPPI. 
Pravitra \etal propose a robust control architecture integrating MPPI with $\mathcal{L}_1$-adaptive control~\cite{pravitra2020}.
Higgins \etal propose an MPPI motion planner that minimizes the cost function including collision risk~\cite{higgins2023}.
Some studies develop a drone navigation framework with MPPI using a cost map~\cite{mohamed2020}, which can handle non-convex and non-differentiable obstacle constraints~\cite{minarik2024}.
However, these works only consider normal drones without wheels.

This paper presents an MPPI-based approach to ground driving and flight locomotion for two-wheeled drones.
To apply MPPI, we overcome the unique challenges of two-wheeled drones, such as the avoidance of unstable landing via a single wheel and the switching between driving and flying modes.
To our knowledge, this is the first study to apply MPPI to two-wheeled drones.
\section{Target Task and Requirements}
\label{sec:task_and_requirements}

\begin{figure}[t]
    \centering
    \includegraphics[width=1\linewidth]{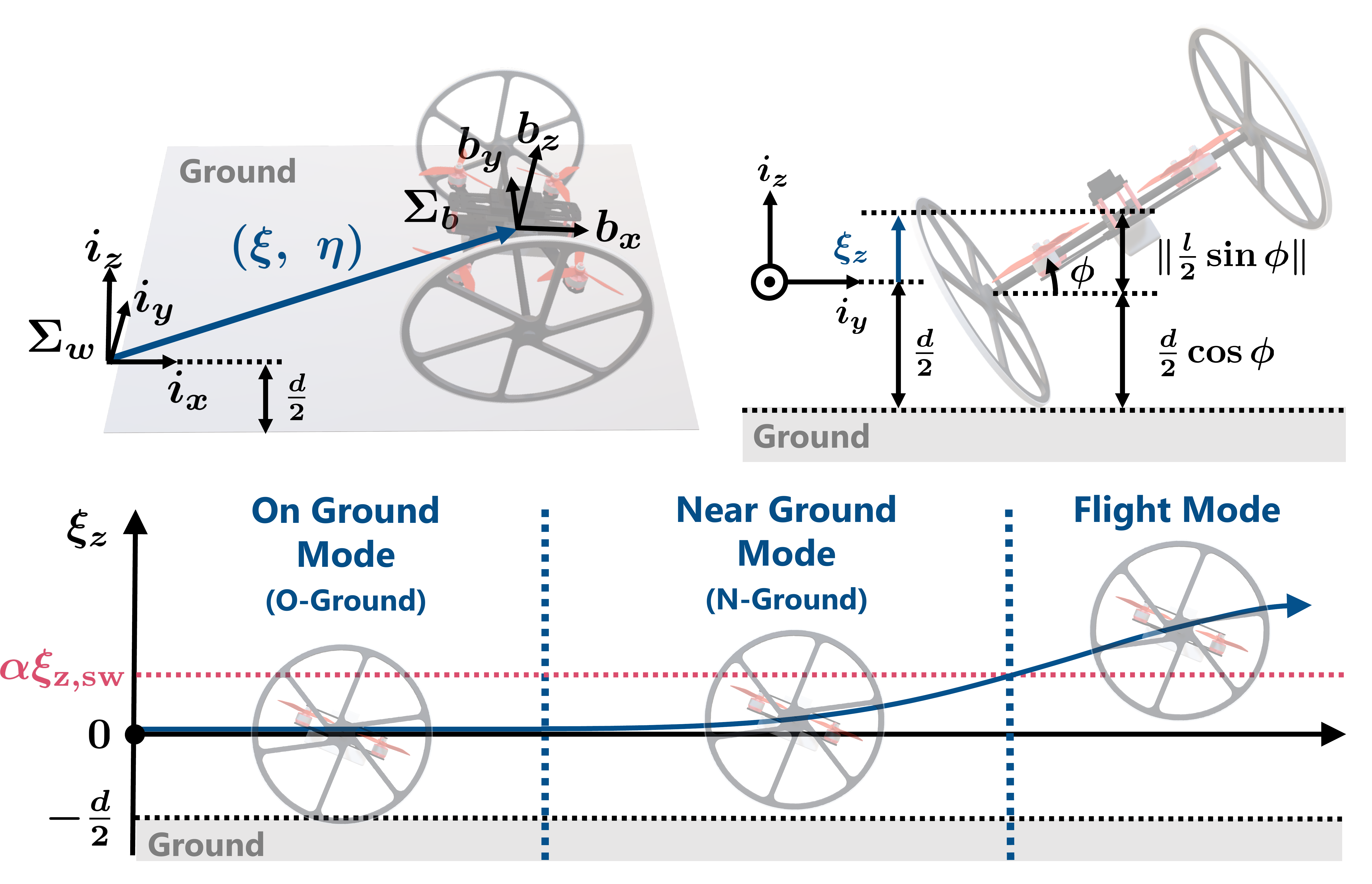}
    \caption{Coordinate frame, geometry, and modes of two-wheeled drone. }
    \label{fig:two_wheel_drone_model}
    \vspace{-5mm}
\end{figure}

\looseness=-1
Two-wheeled drone is a drone with two passive wheels on its side, as shown in Fig.~\ref{fig:two_wheel_drone_model}.
The two-wheeled drone is controlled by four rotors on the drone and can move in the air and on the ground by controlling the thrust distribution.

\looseness=-1
In this work, we aim to achieve point-to-goal navigation with a two-wheeled drone, where the drone reaches a given goal position while avoiding obstacles, as shown in Fig.~\ref{fig:problem_setting}.
We assume that a reference trajectory, \ie, a sequence of 3D positions and velocities, is given by a global trajectory planner, and we focus on motion planning to follow the reference trajectory while avoiding obstacles.
Although the two-wheeled drone can move by flying in the air and driving on the ground, the dynamics become complex and unstable when only one wheel contacts the ground.
Therefore, near the ground, we need to maintain a horizontal attitude so that both wheels land on the ground at the same time.
Furthermore, two-wheeled drones exhibit different dynamics in flight and driving due to contact forces with the ground.
Therefore, we need to consider the multimodality of the dynamics in the design of planner.

\section{Dynamics Modeling of Two-wheeled Drone}
\label{sec:modeling}

\subsection{Derivation of Euler-Lagrange Equations}
To establish the state transition model for the two-wheeled drone in the MPPI framework,
we model the dynamics of two-wheeled drones based on Lagrangian dynamics \cite{murray2017}, \cite{featherstone2008} and normal drone dynamics \cite{nonami2010}.
As shown in Fig.~\ref{fig:two_wheel_drone_model}, let $\Sigma_w = (i_x, i_y, i_z)$ denote the inertial coordinate and $\Sigma_b = (b_x, b_y, b_z)$ denote the body coordinate, which is the center of gravity (CoG) of the two-wheeled drone, where $b_x$ and $b_y$ are on a plane passing through the centers of rotors, $b_y$ has the same direction as the axle, and $b_z$ is vertical to this plane and has the same direction as the thrust generated by the rotors.
Let $\xi =[\xi_x~\xi_y~\xi_z]^\Tp \in \R^3$ be the position of the body coordinates $\Sigma_b$ with respect to the inertial coordinates $\Sigma_w$ .
The attitude of the body coordinates $\Sigma_b$ with respect to the inertial coordinates $\Sigma_w$ is expressed by the ZYX Euler angle $\eta = [\psi~\theta~\phi]^\Tp$.
The rotation matrix of $\Sigma_b$ with respect to $\Sigma_w$ is denoted by $R \in SO(3)$.
Then, the relationship between $\eta$ and $R$ is as follows.
\begin{align*}
    R(\eta)
     & =
    \begin{bmatrix}
        \ssin \theta ~ \scos \psi & \ssin \phi ~ \ssin \theta ~ \scos \psi - \scos \phi \ssin \psi & \scos \phi ~ \ssin \theta ~ \scos \psi + \ssin \phi \ssin \psi \\
        \scos \theta ~ \ssin \psi & \ssin \phi ~ \ssin \theta ~ \ssin \psi + \scos \phi \scos \psi & \scos \phi ~ \ssin \theta ~ \ssin \psi - \ssin \phi \scos \psi \\
        -\ssin \theta             & \ssin \phi ~ \scos \theta                                      & \scos \phi ~ \scos \theta
    \end{bmatrix},
\end{align*}
where
$\ssin ~ \cdot = \sin \cdot,~ \scos ~ \cdot = \cos\cdot$.
From the nature of the rotation matrix, defining $e_x = [1~0~0]^\Tp,e_y = [0~1~0]^\Tp,e_z = [0~0~1]^\Tp \in \R^3$, we have $[b_x~b_y~b_z] = R [e_x~e_y~e_z]$.

Then, the kinematics of a two-wheeled drone is formulated as $\dot{\xi} = R v$ and $\dot{\eta} = \Phi \Omega$, where body velocity and body angular velocity are defined as $v = [v_x~ v_y~ v_z]^\Tp$ and $\Omega = [\Omega_x,~ \Omega_y,~ \Omega_z]^\Tp$, and $\Phi \in \R^{3 \times 3}$ is defined as
\begin{align*}
    \Phi(\eta) =
    \begin{bmatrix}
        0 & -\ssin \phi / \scos \theta & \scos \phi / \scos \theta \\
        0 & \scos \phi                 & -\ssin \theta             \\
        1 & \ssin \phi ~ \stan \theta  & \scos \phi ~ \stan \theta
    \end{bmatrix},
\end{align*}
where $\stan ~ \cdot = \tan \cdot$.
Let $m \in \R$ be the airframe mass, $J = \diag(J_x,J_y,J_z) \in \R^{3 \times 3}$, $d \in \R$ be the wheel diameter, $l \in \R$ be the axle length, $f \in \R$ be the whole thrust generated by the drone in the $b_z$ direction, and $\tau \in \R^3$ be the torque.

In this work, we define the three state-dependent modes in order to derive the dynamics and allow stable navigation near the ground.
As shown in Fig~\ref{fig:two_wheel_drone_model}, the three modes depend on $z$ position $\xi_z$ and are expressed as follows:
\begin{align}
    \mathrm{mode} =
    \begin{cases}
        \mytag{O-Ground} ,~ \mathrm{if} ~ \xi_z = 0                                 \\
        \mytag{N-Ground} ,~ \mathrm{if} ~ 0 < \xi_z \leq \alpha \xi_{z,\mathrm{sw}} \\
        \mytag{Flight} ,~ \mathrm{if} ~ \xi_z > \alpha \xi_{z,\mathrm{sw}}
    \end{cases},
    \label{eq:mode_selection}
\end{align}
where $\alpha \geq 1$ is a clearance parameter for altitude uncertainty and $\xi_{z,\mathrm{sw}}$ is the altitude where both wheels do not contact at any roll angle $\phi$, as shown in Fig.~\ref{fig:two_wheel_drone_model}.
$\xi_{z,\mathrm{sw}}$ is expressed as follows based on the geometrical relationship:
\begin{align*}
    \xi_{z,\mathrm{sw}} = \max_{\phi \in (-\pi/2,~\pi/2)} \left[ \frac{d}{2} \cos \phi + \|\frac{l}{2} \sin \phi\| - \frac{d}{2} \right].
\end{align*}

\looseness=-1
When the two-wheeled drone drives, \ie, $\mathrm{mode} = \mytag{O-Ground}$, it is constrained on the ground.
We assume that the ground is a plane vertical to $i_z$ at $-d/2$ in the $i_z$ direction, as shown in Fig.~\ref{fig:two_wheel_drone_model}.
Two-wheeled drone on the ground satisfies the following three constraints:
1) The distance between the CoG and the ground is constant, which can be expressed as
\begin{align}
     & h_1 = \xi_z =
    \begin{bmatrix}
        0 & 0 & 1
    \end{bmatrix}
    \xi = 0. \label{eq:constraint1}
\end{align}
2) The wheels do not skid on the ground as
\begin{align}
     & h_2 =
    v_y =
    \begin{bmatrix}
        -\sin \psi & \cos \psi & 0
    \end{bmatrix}
    \dot{\xi} = 0. \label{eq:constraint2}
\end{align}
3) The drone does not rotate in the roll direction as
\begin{align}
     & h_3 =
    \begin{bmatrix}
        0 & 0 & 1
    \end{bmatrix}
    \eta = 0. \label{eq:constraint3}
\end{align}
These constraints can be summarized in Pfaffian form as $A_\xi \dot{\xi} = 0$ and $A_\eta \dot{\eta} = 0$ in the translation and rotation directions, respectively, where
\begin{align*}
     & A_\xi   =
    \begin{bmatrix}
        0          & 0         & 1 \\
        -\sin \psi & \cos \psi & 0
    \end{bmatrix},~
    A_\eta  =
    \begin{bmatrix}
        0 & 0 & 1
    \end{bmatrix}.
\end{align*}
Thus, the constraints $h_i ~ (i = 1,2,3)$ due to contact with the ground can be expressed as a Lagrange multiplier $\lambda_i$ when driving on the ground \cite{murray2017}.
On the other hand, when the two-wheeled drone is in flight, \ie, $\mathrm{mode} \in \{\mytag{N-Ground},\mytag{Flight}\}$, $h_1 > 0$ and these three constraints in \eqref{eq:constraint1}, \eqref{eq:constraint2}, and \eqref{eq:constraint3} are not satisfied since the wheels are not in contact with the ground.
Note that we do not model the rotational dynamics in landing with one wheel.
This is because, when the two-wheeled drone flies near the ground, $\mathrm{mode} = \mytag{N-Ground}$, the roll angle $\phi$ is assumed to be controlled zero, and the drone maintains a horizontal attitude.

Finally, we derive the Euler-Lagrange equations of motion in the translational and rotational directions of the two-wheeled drone when both driving and flying~\cite{nonami2010, murray2017}, as follows:
\begin{subequations}
    \label{eq:dynamics}
    \begin{align}
        \label{eq:transrational_dynamics}
         & m \ddot{\xi} + mge_z + A_{\xi}^\Tp
        \begin{bmatrix}
            \lambda_1 \\
            \lambda_2
        \end{bmatrix}
        = f R e_z,                                                                 \\
        \label{eq:rotational_dynamics}
         & M \ddot{\eta} + C \dot{\eta}  + A_{\eta}^\Tp \lambda_3 = \Psi^\Tp \tau,
    \end{align}
\end{subequations}
\begin{align}
     &
    \label{eq:constraint_force}
    \begin{bmatrix}
        \lambda_1 \\
        \lambda_2 \\
        \lambda_3
    \end{bmatrix}
    =
    \begin{cases}
        \begin{bmatrix}
            m \dot{A}_\xi \dot{\xi} + A_\xi \big(f R e_z - m g e_z \big) \\
            \big(A_\eta M^{-1} A_\eta \big)^{-1} A_\eta M^{-1} \big(\Psi^\Tp \tau  ~ - C \dot{\eta}\big)
        \end{bmatrix} \\
        ~,~ \mathrm{if ~ } \mathrm{mode} = \mytag{O-Ground}                                         \\
        [0~0~0]^\Tp                                                                                 \\ ~,~ \mathrm{if} ~ \mathrm{mode} \in \{\mytag{N-Ground},\mytag{Flight}\}
    \end{cases}, \\
     & \Psi(\eta) = \Phi^{-1}, \; M(\eta) = \Psi^\Tp J \Psi,                              \\
     & C(\eta,\dot{\eta}) = \Psi^\Tp J \dot{\Psi} + \Psi^\Tp \sk(\Psi \dot{\eta}) J \Psi,
\end{align}
Note that $\lambda_1 \leq 0$ is satisfied since it is a reaction force from the ground.

\subsection{Discontinuous Impulse at Ground Contact}
At the moment of contact with the ground, the state of the two-wheeled drone changes discontinuously due to ground repulsion and the generation of constraints.
Based on the generation of the constraint \eqref{eq:constraint3}, the attitude and angular velocity after the contact $\eta^+,\dot{\eta}^+$ are expressed as
\begin{align}
    \label{eq:init_eta}
    \eta^+ = T_3 \eta^-,\dot{\eta}^+ = T_3 \dot{\eta}^-,
\end{align}
where $T_3 = \diag(1,1,0)$ and $\eta^-,\dot{\eta}^-$ are attitude and angular velocity before the contact.
The ground repulsion with regard to rotation is ignored, since we assume that both wheels are in almost horizontal and simultaneous contact.

Furthermore, we derive the velocity after contact with the ground $\dot{\xi}^+$ from the velocity and attitude before contact $\dot{\xi}^-,\eta^-$.
The velocity after ground contact, considering only the repulsion from the ground whose normal vector is $e_z$, is expressed as $T_1 \dot{\xi}^-$, using $T_1 = \diag(1,1,-e)$ and the repulsion coefficient $e \in \R$ \cite{featherstone2008}.
Considering the generation of nonholonomic constraints \eqref{eq:constraint2}, the body velocity after contact is expressed as $T_2 R^\Tp(T_3 \eta^-) \dot{\xi}^-$, where $T_2 = \diag(1,0,1)$.
Thus, the velocity after contact is expressed as:
\begin{align}
    \label{eq:init_dot_xi}
    \dot{\xi}^+ = T_1 R(T_3 \eta^-) T_2 R^\Tp(T_3 \eta^-) \dot{\xi}^-.
\end{align}

\section{Two-wheeled Drone Navigation using MPPI}
\label{sec:controller_design}

\begin{figure}[t]
    \centering
    \includegraphics[width=0.9\linewidth]{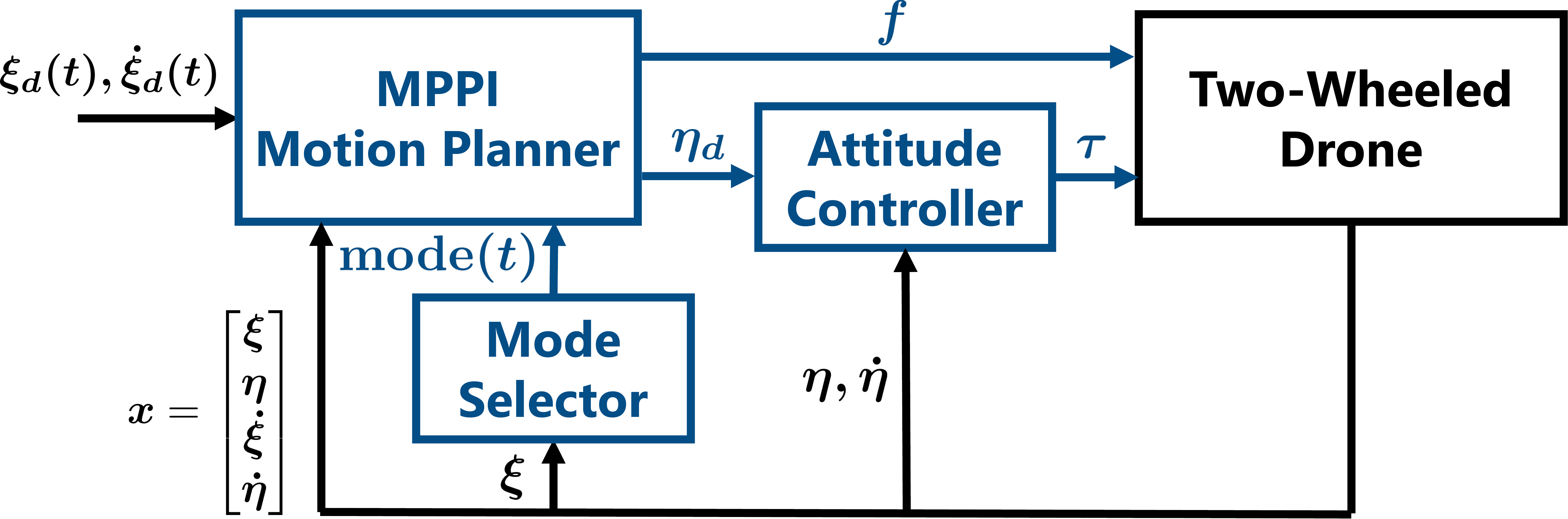}
    \caption{Overview of the navigation system for the two-wheeled drone.} 
    \label{fig:block_diagram}
    \vspace{-5mm}
\end{figure}

\looseness=-1
In this paper, we propose a two-wheeled drone navigation system that allows a two-wheeled drone to navigate to a given goal position while switching the modes described in \eqref{eq:mode_selection}.
The proposed framework, as illustrated in Fig.~\ref{fig:block_diagram}, consists of three main components: a \emph{Mode Selector}, a \emph{MPPI-based motion planner} and an \emph{attitude controller}.
The mode selector determines the current mode at time $t$, \ie, $\mathrm{mode}(t)$, based on \eqref{eq:mode_selection} .
Depending on the current mode, the MPPI-based motion planner generates thrust $f$ and reference attitude $\eta_d$ to follow a given time-varying reference position $\xi_d(t)$ and velocity $\dot{\xi}_d(t)$ while avoiding obstacles.
Then, the attitude controller generates the torque $\tau$ to make the drone's attitude $\eta$ follow the reference attitude $\eta_d$, and $[f ~ \tau]^\Tp$ is sent to the two-wheeled drone.

\subsection{Motion planning based on MPPI}
\label{sec:position_controller}
MPPI is a sampling-based MPC that solves a finite-horizon optimal control problem by stochastic optimization.
In this work, compared to the normal MPPI-based planner, the dynamics and control requirements change depending on the mode.
Therefore, we design a mode-specific input space and introduce the auxiliary controller to extend MPPI.

\subsubsection{Design of Mode-specific Input Space}
As described in Section \ref{sec:task_and_requirements}, the two-wheeled drone becomes unstable when only a single wheel is landed.
Therefore, it is necessary to maintain the roll angle of the drone always at zero in \mytag{N-Ground} mode.
On the other hand, in \mytag{Flight} mode, it is preferable to lock the yaw angle (set to zero) because it does not affect translational motion, and instead, make full use of the roll degree of freedom.
Based on these requirements, we switch the input space of MPPI at time $t$ according to the mode determined by the mode selector as follows:
\begin{align}
    \label{eq:input_space}
    & u= 
    \begin{cases}
        \begin{bmatrix}
            f & \psi_d & \theta_d & 0
        \end{bmatrix}^T     \\
        ~,~ \mathrm{if ~} \mathrm{mode(t)} \in \{\mytag{O-Ground},\mytag{N-Ground}\}\\
        \begin{bmatrix}
            f & 0 & \theta_d & \phi_d
        \end{bmatrix}^T 
        ~,~ \mathrm{if ~} \mathrm{mode(t)} = \mytag{Flight}
    \end{cases}.
\end{align}
Similarly to this approach, some works have shown that it is possible to adapt to various situations by switching input spaces depending on the situation~\cite{aoki2024}.

\subsubsection{MPPI Formulation for Two-Wheeled Drone}

MPPI solves the following stochastic optimization problem at time $t$ to find the optimal control input for a finite horizon $T$:
\begin{align}
    & {{U}}^{*}=\underset{{U}}{\argmin}\,
    \mathbb{E}\left[c_{\mathrm{term}}(x_{T}) + \sum_{j = 0}^{T-1} \mathcal{L}(x_j,u_j) \right], \\
    & \mathrm{s.t.} ~ {x}_{j+1} = F({x}_{j},{u}_{j}),\; x_0 = x(t), 
    \label{eq:optimization_problem}
\end{align}
where $j = \{0, \dots, T-1\}$, $x = [\xi^\Tp ~ \eta^\Tp ~ \dot{\xi}^\Tp ~ \dot{\eta}^\Tp]$, $U = [u_0, \dots, u_{T-1}]$, $u$ is the control input in \eqref{eq:input_space}, 
$\mathcal{L}(x_j,u_j) = c({x}_{j}, {u}_{j}) + \frac{\lambda}{2}{{u}}_{j}^{\rm{T}}\Sigma^{-1}{{u}}_{j}$, $c$ is the running cost function, $c_{\mathrm{term}}$ is the terminal cost function, $\lambda$ is the temperature parameter, $\Sigma$ is the variance, and $F$ is the state transition model.

\textbf{{State Transition Model}:} 
We formulate the state transition model $F$ in \eqref{eq:optimization_problem}.
First, except for the moment of contact, the state transition model for both driving and flight is derived from the translational Euler-Lagrange equations \eqref{eq:transrational_dynamics} discretized by the forward Eulerian method, as follows:
\begin{align*}
    x_{j+1} &= F_{\mathrm{eul}}(x_j,u) \\
    &= 
    \begin{bmatrix}
        \xi_j + \dot{\xi}_j \Delta t   \\
        \eta_d                         \\
        \dot{\xi}_j + \left(\frac{1}{m} f R_j e_z - g e_z  - A_{\xi}^\Tp
        \begin{bmatrix}
            \lambda_1 \\
            \lambda_2
        \end{bmatrix} \right) \Delta t \\
        (\eta_d - \eta_j)/ \Delta t
    \end{bmatrix},
\end{align*}
where $F_{\mathrm{eul}}$ includes the dynamics of both steady ground contact and flight.
The contact force is expressed by the Lagrange multipliers \eqref{eq:constraint_force}, assuming that the initial velocity in the constrained direction is zero.
However, the impulsive contact force that occurs at the moment of contact is not considered.
Thus, we model the dynamics at the moment of contact based on the \eqref{eq:init_eta}, \eqref{eq:init_dot_xi} as follows:
\begin{align*}
    x_{j+1} = F_{\mathrm{cnt}}(x_j,u) = 
    \begin{bmatrix}
        \xi_j + \dot{\xi}_j \bar{\Delta t}      \\
        T_3 \eta_d                     \\
        T_1 R(T_3 \eta_j) T_2 R^\Tp(T_3 \eta_j) \dot{\xi}_j           \\
        T_3 (\eta_d - \eta_j)/\bar{\Delta t} \\
    \end{bmatrix},
\end{align*}
where $\bar{\Delta t} = -\xi_{z,j}/\dot{\xi}_{z,j}$ is the time step for the altitude to be exactly zero in the next step.
In summary, the full-state transition model $F$ is expressed as follows:
\begin{align}
    x_{j+1} &= F(x_j,u) \label{eq:transition_model} \\
    &=
   \begin{cases}
       F_{\mathrm{cnt}}(x_j,u),                            
        ~ \mathrm{if ~ } \xi_{z,j} > 0 ~ \& ~ \xi_{z,j} + \dot{\xi}_{z,j} \Delta t \leq 0\\
       F_{\mathrm{eul}}(x_j,u)
       ,~\mathrm{otherwise}\notag
   \end{cases},
\end{align}
where we detect the moment of contact by checking the altitude of the next step $\xi_{z,j} + \dot{\xi}_{z,j} \Delta t$ using constant sample time $\Delta t$ when the altitude of the current step is higher than zero $\xi_{z,j} > 0$.
Note that the rotational dynamics $\eqref{eq:rotational_dynamics}$ is ignored in the state transition model, assuming that the attitude $\eta$ is consistent with the reference attitude $\eta_d$.

\textbf{{Cost Function}: }
We design the running and terminal cost functions to follow the reference trajectory $\xi_d(t)$ and $\dot{\xi}_d(t)$, and avoid obstacles as follows:
\begin{align*}
    & c(x,u) = \| \xi_e \|^2_{W_{\xi}} + \| \dot{\xi}_e \|^2_{W_{\dot{\xi}}} + \| u \|^2_{W_u} + W_{\mathrm{obs}} \mathds{1}_\mathrm{obs}, \\
    & c_{\mathrm{term}}(x) = \| \xi_e \|^2_{W_{\xi,\mathrm{term}}} + \| \dot{\xi}_e \|^2_{W_{\dot{\xi},\mathrm{term}}},
\end{align*}
where the state error $\xi_e = \xi - \xi_d(t),\dot{\xi}_e = \dot{\xi} - \dot{\xi}_d(t)$, $W_{\xi}, W_{\dot{\xi}}, W_{\xi,\mathrm{term}}, W_{\dot{\xi},\mathrm{term}} \in \R^{3 \times 3}$, $W_u \in \R^{4 \times 4}$ is a semi-positive definite weight matrix, and $\| \cdot \|_P$ is a weighted Euclidean inner product $\| u \|_P^2 = u^\Tp P u$, 
$W_{\mathrm{obs}} \in \R$ as a collision weight and $\mathds{1}_\mathrm{obs}$ is a non-differentiable indicator function that takes the value 1 in case of obstacle collision and 0 otherwise.

\textbf{{MPPI Solver}: }
MPPI can solve the optimization problem \eqref{eq:optimization_problem} through an importance sampling technique as follows:
\begin{align}
    &u^{*}_{j}(t) = \sum_{k=1}^{K}  w_k u_{j,k}, \label{eq:solve_mppi} \\ 
    &u_{j,k} = \hat{u}_{j} + \epsilon_{j,k},\; \epsilon_{j,k} \sim \mathcal{N}(0,\Sigma), \\
    & w_k = \mathrm{Softmax} \Big( -\lambda^{-1} \big( c_{\rm{term}}(x_{T,k}) + \sum_{j=0}^{T-1} \mathcal{L}({x}_{j,k}, {u}_{j,k}) \big) \Big), \notag
\end{align}
where $\hat{U}(t) = [\hat{u}_0, \dots, \hat{u}_{T-1}]$ is the prior input sequence, $k = \{1, \dots, K\}$, $K$ is the number of samples.
The updated solution $u^{*}$ minimizes the Kullback-Leibler divergence with the optimal control distribution characterized by the Boltzmann distribution.

\subsubsection{Prior input sequence by auxiliary controller}
\label{sec:initialization}

In the standard MPPI, \eqref{eq:solve_mppi} is solved based on the warm start manner, where the prior input sequence $\hat{U}(t)$ is used as the optimal solution of the previous step $U^*(t-1)$.
The prior input sequence acts as a regularization term that penalizes deviations from $U^*(t-1)$.
Although this penalty is effective for smoothing the control input, it has an adverse effect when a sudden input change is required~\cite{trevisan2024,honda2024stein}.
As the two-wheeled drone requires a sudden input change to transition from driving mode to flying mode, we mix the samples with an auxiliary controller-based input sequence that includes gravity compensation.
The prior input sequence, in this work, is proposed as follows:
\begin{align*}
    \hat{u}_{j,k}(t) =
    \begin{cases}
        u_j^{\mathrm{aux}}(t)     & ,~ \mathrm{if} ~ 1 \leq k \leq K_{\mathrm{aux}} \\
        u_j ^{*}(t-1) & ,~ \mathrm{otherwise}
    \end{cases},
\end{align*}
where $u^{*}$ is the optimal solution of the previous time $t-1$, $K_{\mathrm{aux}}$ is the number of samples for the auxiliary controller-based input sequence, and $u_j^{\mathrm{aux}}$ is the auxiliary controller-based input sequence.
$u_j^{\mathrm{aux}}$ is calculated over the horizon from the state transition model \eqref{eq:transition_model} and the control method of normal drones ~\cite{nonami2010} as follows:
\begin{align*}
     & u_j^{\mathrm{aux}}(t) =
    \begin{bmatrix}
        m \sqrt{\mu^2_x + \mu^2_y + (\mu_z + g)^2}                                         \\
        \arcsin \left( -\frac{\mu_y}{\sqrt{\mu^2_x + \mu^2_y + (\mu_z + g)^2}} \right) \\
        \arctan \left(\frac{\mu_x}{\mu_z + g} \right)                                      \\
        0
    \end{bmatrix},                             \\
     & [\mu_x~ \mu_y~ \mu_z]^\Tp = -K_{\xi} (\xi_j - \xi_{d,j}(t)) ~ -K_{\dot{\xi}} (\dot{\xi}_j - \dot{\xi}_{d,j}(t)),
\end{align*}
where $K_{\xi},K_{\dot{\xi}} \in \R^{3 \times 3}$ is the proportional and differential gain.

\subsection{Attitude Controller}
\label{sec:attitude_controller}

To control the attitude of the two-wheeled drone to follow the reference attitude $\eta_d$ calculated by the MPPI-based motion planner,
the torque input $\tau$ is computed by the attitude control law ~\cite{nonami2010} as follows:
\begin{align*}
    & \tau = J \Psi (-K_\eta e_\eta -K_{\dot{\eta}} e_{\dot{\eta}}) + \Psi^{-1} C \dot{\eta},\\
    & e_\eta = \eta - \eta_d,~ e_{\dot{\eta}} = \dot{\eta} - \dot{\eta}_d,
\end{align*}
where $K_\eta,~ K_{\dot{\eta}} \in \R^{3 \times 3}$ is the proportional and differential gain.

\section{Experiments}
\label{sec:simulation}

In this section, we demonstrate that our proposed method achieves the point-to-goal navigation described in Section~\ref{sec:task_and_requirements} on numerical simulations.

\subsection{Simulation Setups}

\looseness=-1
In the simulated environment with three cylinder obstacles, as shown in Fig.~\ref{fig:problem_setting}, the two-wheeled drone must navigate to the goal position $\xi_g = [3~0.5~0]^\Tp$ from the initial settings: $\xi_0 = [0~0~0]^\Tp$, $\dot{\xi}_0 = [0~0~0]^\Tp$, $\eta_0 = [0~0~0]^\Tp$ and $\dot{\eta}_0 = [0~0~0]^\Tp$.
Three cylindrical obstacles, each with a radius of \(0.05\,\unit{m}\), are placed in \((2.0,\,0.0,\,0.14)\), \((0.6,\,0.15,\,0.0)\), and \((1.6,\,0.05,\,0.0)\), and are oriented in the \(i_y\), \(i_z\), and \(i_z\) directions, respectively.
We assume that the drone can fully observe the obstacles.

\looseness=-1
We simulate the two-wheeled drone based on the dynamics described in Section~\ref{sec:modeling} with the control rate of $\Delta t = 0.02\,\unit{s}$.
For physical parameters, the drone mass is $m = 0.938 \unit{kg}$, the inertia tensor is $J = \diag (0.00933, 0.00285, 0.01130)\unit{kg.m^2}$, the wheel diameter is $d = \qty{0.28}{m}$, the axle length is $l = 0.35 \unit{m}$, and the repulsion coefficient with the ground is $e = 0.1$.
The collision with the obstacle is evaluated based on the position of CoG $\xi$ and the area offset from the obstacle area by $\sqrt{d^2+l^2}/2$.

\subsection{Implementation Details}
\looseness=-1
As described in Section~\ref{sec:task_and_requirements}, we assume the time-varying reference trajectory.
The reference trajectory is given as a trapezoidal velocity waveform with a slope of $0.5 \unit{m/s^2}$ and a height of $0.5 \unit{m/s}$, connecting $\xi_0$ and $\xi_g$ with a straight line.

We set the weight parameters of the cost function in MPPI as follows: $W_{\xi} = 10^4 \diag(0.9,1.2,0.3)$, $W_{\dot{\xi}} = 10^4 \diag(0.9,1.2,0.15)$, $W_{\xi,\mathrm{term}} = 10^4 \diag(0.75,1,0.275)$, $W_{\dot{\xi},\mathrm{term}}  = 10^4 \diag(0.25,0.25,0.125)$, $W_u = \diag(3.2,1.6,1.6,1.6)$, and $W_{\mathrm{obs}} = 10^6$.
The switching altitude is $ \alpha \xi_{z,\mathrm{sw}} = 0.1261 \unit{m}$, the total number of samples is $K = 1500$, the number of samples for the auxiliary controller is $K_\mathrm{aux} = 300$, the temperature parameter $\lambda = 10$, the gain matrix of the auxiliary controller $K_\xi = \diag(1,1,1),~K_{\dot{\xi}} = \diag(1,1,1)$, the proportional and derivative gain of the attitude control law $K_\eta = \diag(20,20,20 ),~ K_{\dot{\eta}} = \diag(10,10,10)$ and the variance $\Sigma$ is $\diag(2.25,0.03,0.03,0.03)$.

\subsection{Simulation Results}
\begin{figure}[t]
    \begin{minipage}[t]{0.49\linewidth}
        \centering
        \includegraphics[width=1\linewidth]{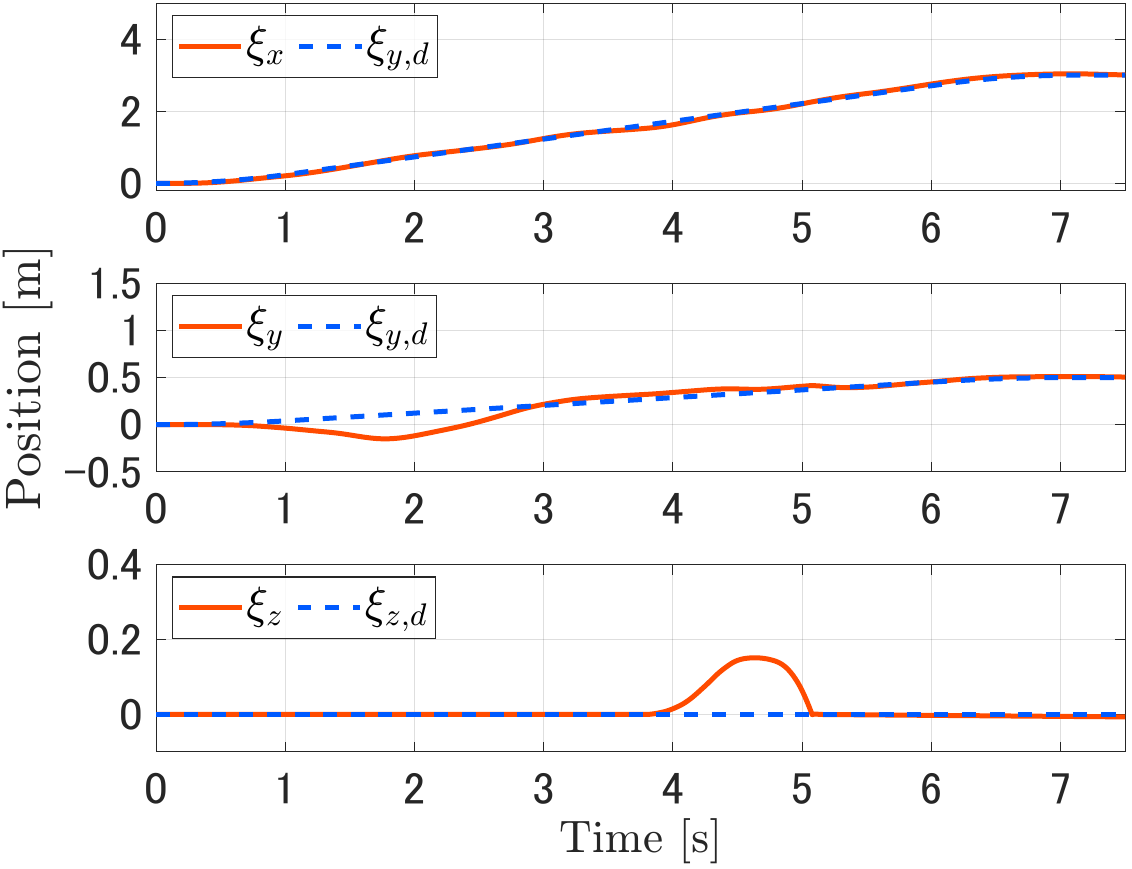}
        \label{fig:sim_position}
    \end{minipage}
    \hfill
    \begin{minipage}[t]{0.49\linewidth}
        \centering
        \includegraphics[width=1\linewidth]{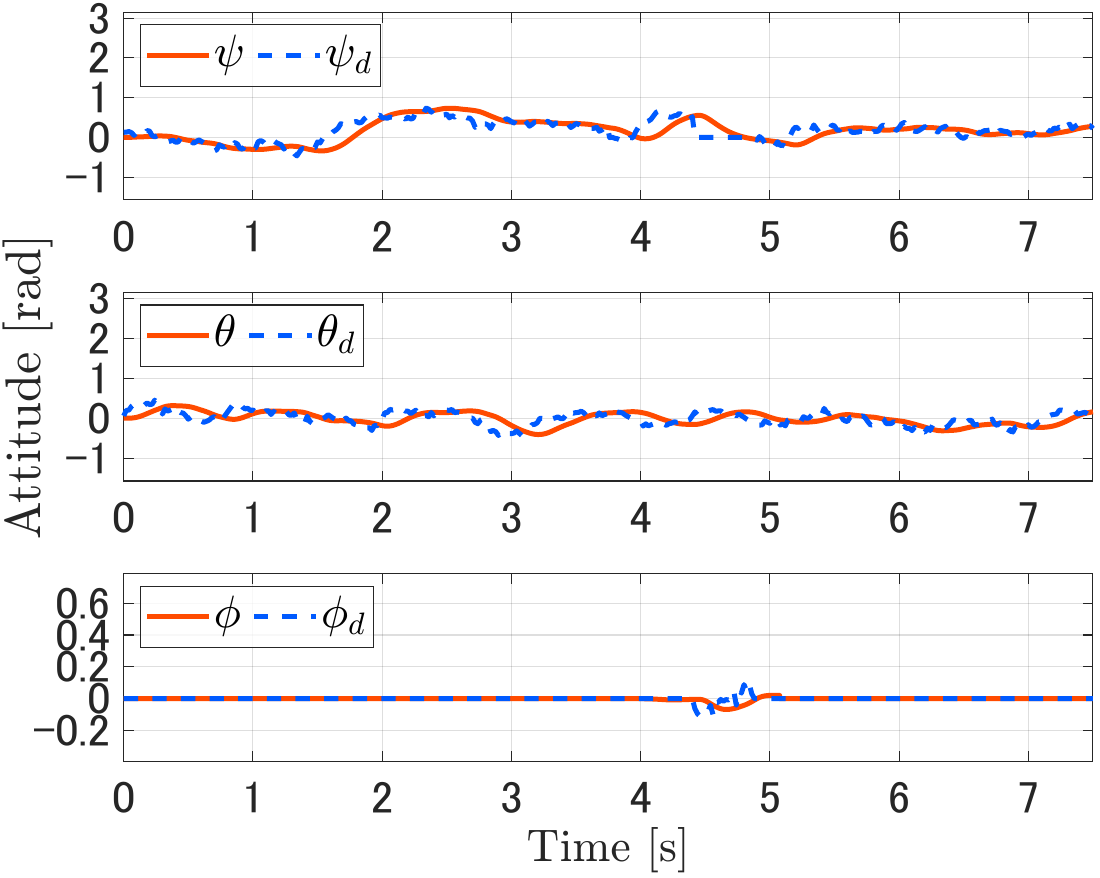}
        \label{fig:sim_attitude}
    \end{minipage}
    \caption{Simulation results of position and attitude}
    \label{fig:sim_position_attitude}
    \vspace{-3mm}
\end{figure}
\begin{figure}[t]
    \centering
    \includegraphics[width=0.85\linewidth]{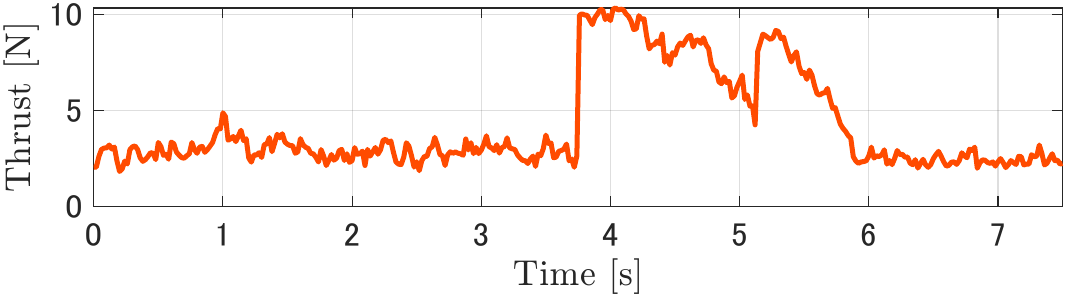}
    \caption{Simulation results of thrust}
    \label{fig:sim_thrust}
    \vspace{-5mm}
\end{figure}

\begin{figure*}[t]
    \centering
    \begin{minipage}[b]{\linewidth}
        \centering
        \includegraphics[width=0.95\linewidth]{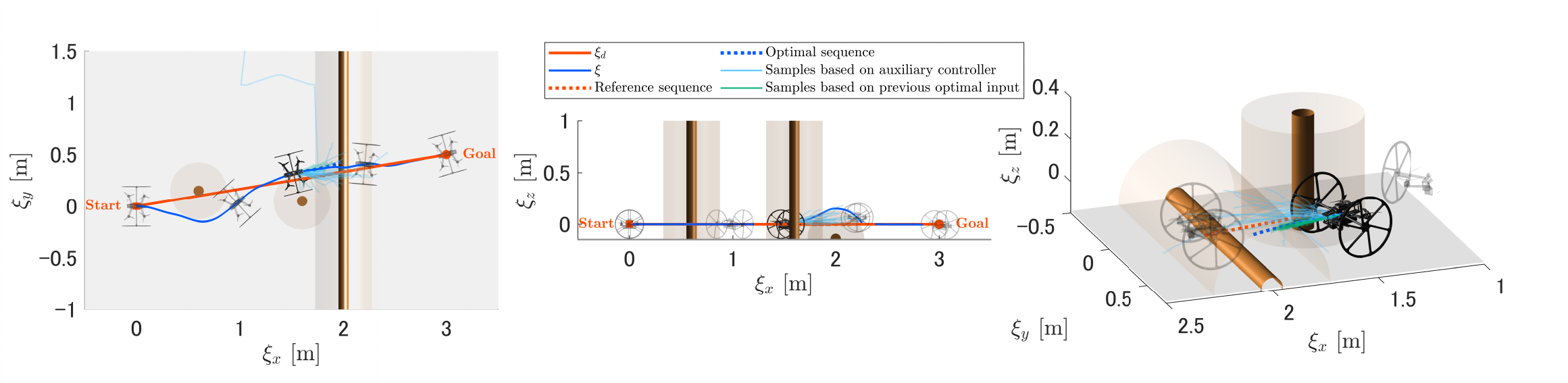}
        \subcaption{A simulation result with the proposed MPPI-based motion planner.}
        \label{fig:sim_trajectory_ours}
    \end{minipage}    
    \begin{minipage}[b]{\linewidth}
        \centering
        \includegraphics[width=0.95\linewidth]{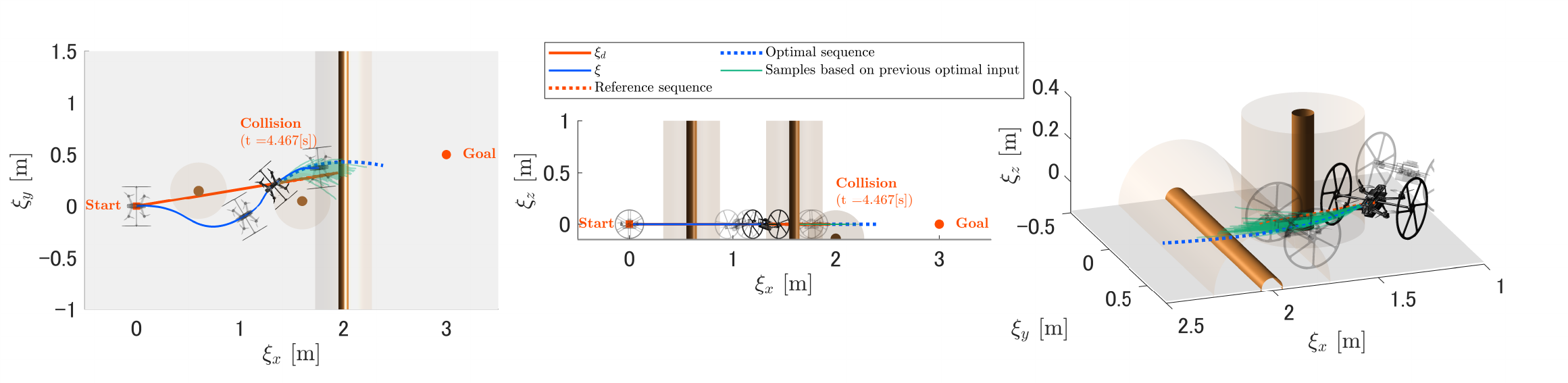}
        \subcaption{A simulation result with the MPPI-based motion planner without the auxiliary controller.}
        \label{fig:sim_trajectory_baseline}
    \end{minipage}
    \caption{Comparison of the trajectories with and without the proposed auxiliary controller.}
    \label{fig:sim_trajectory}
    \vspace{-5mm}
\end{figure*}

Figure~\ref{fig:sim_position_attitude} shows the position and attitude plots of the two-wheeled drone during navigation, where the dashed lines represent the reference trajectory, and the actual trajectory of the drone is shown as solid lines.
Our proposed method successfully follows the reference trajectory while avoiding obstacles mainly by driving on the ground, and the drone reaches the goal position.
When avoiding the obstacle lying on the ground around $t=4.0 \unit{s}$, we can see that the drone temporarily increases the thrust, as shown in Fig.~\ref{fig:sim_thrust}, and avoids the obstacles by flying in the air.

Figure~\ref{fig:sim_trajectory} compares the trajectories with and without the proposed auxiliary controller of MPPI as described in Section~\ref{sec:initialization}.
While the MPPI without the auxiliary controller fails to avoid the obstacles by flying, and collides with the obstacle, our proposed method with the auxiliary controller successfully avoids the obstacles and reaches the goal position.
This result demonstrates that the auxiliary controller is essential for the MPPI-based motion planner to overcome the mode switching from driving to flying.

\section{Conclusion}
This paper proposes a novel MPPI-based navigation framework for a two-wheeled drone.
First, we derive the dynamics of the two-wheeled drone, considering both steady and impulsive contact with the ground.
We then propose an extension of MPPI with a switched input space and an auxiliary controller to handle mode switching, while incorporating non-differentiable obstacle constraints and dynamics into the optimization problem.
As a result, our framework enables navigation that dynamically switches between driving on the ground and flying in the air to avoid obstacles.

\bibliographystyle{IEEEtran}

\end{document}